\documentclass[final,3p,times,twocolumn]{elsarticle}

\usepackage{multirow}
\usepackage{amssymb}
\usepackage{booktabs}
\usepackage{tabularx}
\usepackage{algorithm}
\usepackage{algorithmic}
\usepackage{graphicx}
\usepackage{amsmath}

\journal{Neurocomputing}

\begin{document}

\begin{frontmatter}


\author{Sheng Yi\fnref{label1}}
\author{Xi Li\fnref{label1}}


\title{WSOD with PSNet and Box Regression}

\address[label1]{Department of Electronic Engineering, Tsinghua University}

\begin{abstract}
Weakly supervised object detection(WSOD) task uses only image-level annotations to train object detection task. WSOD does not require time-consuming instance-level annotations, so the study of this task has attracted more and more attention. Previous weakly supervised object detection methods iteratively update detectors and pseudo-labels, or use feature-based mask-out methods. Most of these methods do not generate complete and accurate proposals, often only the most discriminative parts of the object, or too many background areas. To solve this problem, we added the box regression module to the weakly supervised object detection network and proposed a proposal scoring network (PSNet) to supervise it. The box regression module modifies proposal to improve the IoU of proposal and ground truth. PSNet scores the proposal output from the box regression network and utilize the score to improve the box regression module. In addition, we take advantage of the PRS algorithm for generating a more accurate pseudo label to train the box regression module. Using these methods, we train the detector on the PASCAL VOC 2007 and 2012 and obtain significantly improved results.
\end{abstract}





\end{frontmatter}


\section{Introduction}
\label{sec1}
The object detection task is to find the objects belonging to specified classes and their locations in the images. Benefiting from the rapid development of deep learning in recent years, the fully supervised object detection task has made significant progress. However, the fully supervised task requires instance-level annotation for training, which costs a lot of time and resources. In fact, unlabeled or image labeled datasets cannot be effectively used by the fully supervised method. On the other hand, image-level annotated datasets are easy to generate, and can even be automatically generated by web search engines. In order to effectively utilize these readily available datasets, we focus on weakly supervised object detection(WSOD) task. The WSOD task only takes the image-level annotations to train the instance-level object detection network, which is different from the fully supervised object detection task. 

\begin{figure}[t]
	\begin{center}
		\includegraphics[width=0.8\linewidth]{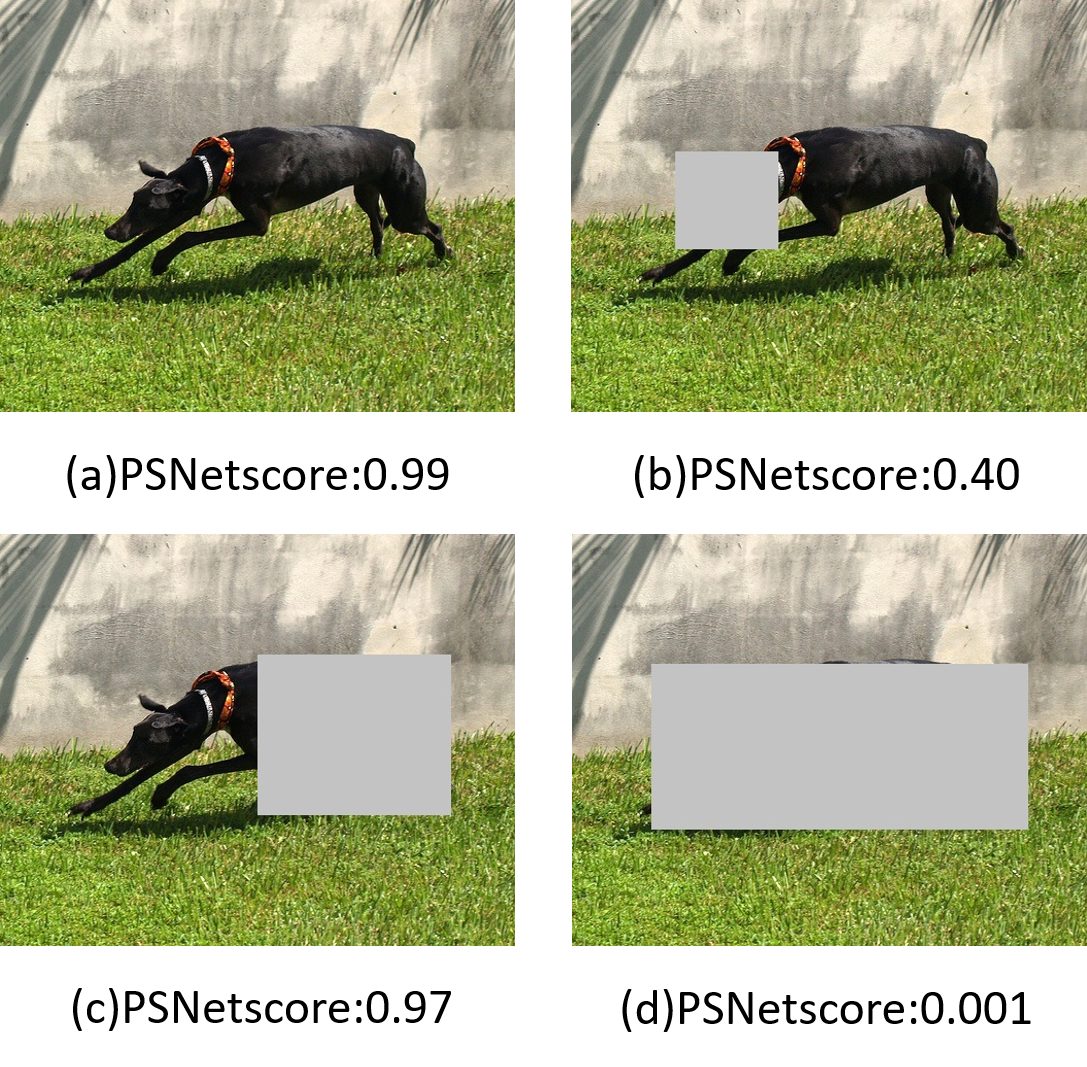}
	\end{center}
	\caption{Examples of PSNet outputs: (a) a dog without proposal occlusion, (b) a dog whose head is occluded by the proposal box, (c) a dog that proposal covers part of the body, and (d) proposal completely cover the entire dog. If proposal does not completely cover the entire dog, PSNet gives a high score. If proposal completely cover the entire dog, PSNet gives a low score.}
	\label{fig:figure1}
\end{figure}

There are three main methods for weakly supervised object detection: The first is to iteratively update the detector and pseudo labels from inaccurate pseudo labels; The second is to construct an end-to-end network that can take image-level annotations as supervision to train this object detection network. The third two-stage method is that taking algorithm to optimize pseudo labels from other WSOD network and training a fully supervised object detection network. In addition, according to different methods of proposing proposals, each of the above methods can be divided into two classes: one is to propose proposals based on feature map that predicts the probability of each pixel belonging to each classes, and then get the possible instances and their locations in the image; The second is detector-based method that uses a trained detector to identify multiple proposals and determine whether each proposal belongs to a specific object class or not. Comparing the effects of these methods, the end-to-end detector-based approach perform well, and our work follows this series of methods.

The earliest end-to-end detector-based WSOD network is WSDDN~\cite{bilen2016weakly}, which trains a two-streams network to predict the classification accuracy of each proposal and their contributions to each class. The results of the two streams are combined to get the image classification score, so the WSDDN can take advantage of the image-label annotations for training. Subsequent other work aims to improve the performance of this network, like adding more classification streams, using clustering method, adding fully supervised module, and so on. The end-to-end detector-based approach has two drawbacks: one is that context information cannot be fully used to classify the proposal; The second is that the most discriminative parts of the object may be detected instead of the entire object.

\begin{figure*}
	\begin{center}
		\includegraphics[width=0.9\linewidth]{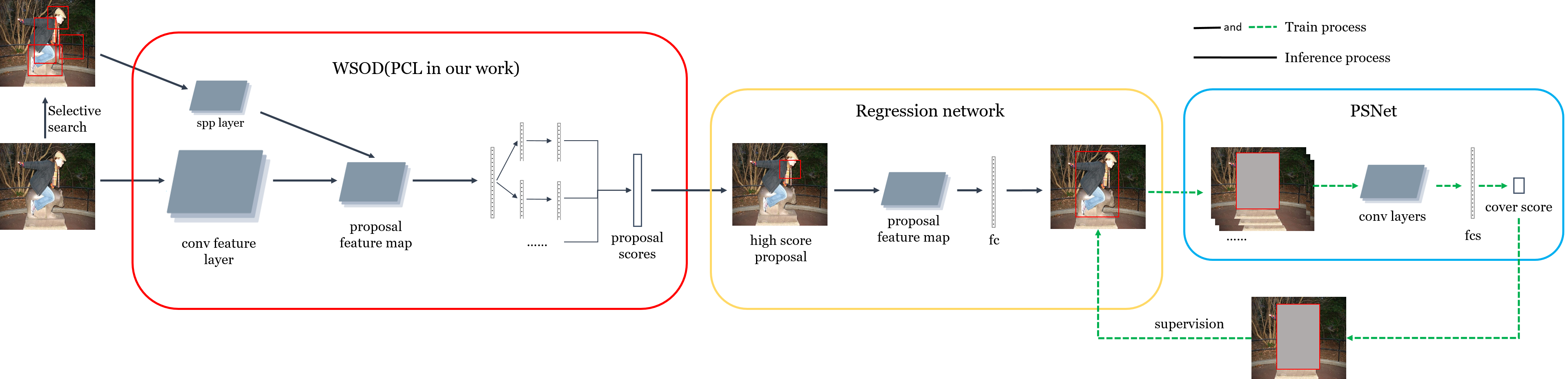}
	\end{center}
	\caption{The figure above is architecture of our work: In the red box is the backbone WSOD network, and in our work it is the PCL network. In the yellow box is box regression network. In the blue box is our PSNet. In training process: (1) The image is first tested with the PCL network to generate 4096-dimension feature of proposals extracted by the convolution module and the corresponding detection scores. (2) Then the box regression network makes a regression correction to the proposals. (3) The output proposal is given to the PSNet to predict the proposal completeness score. The PSNet network then uses the PRS algorithm to find the optimal proposal, which is used as a pseudo label training box regression network. In inference process, the PSNet does not work.}
	\label{fig:figure2}
\end{figure*}

In order to make full use of the context information of proposal and avoid finding only the most discriminative part, we design a new network structure that add a box regression branch to a traditional WSOD network. In the previous WSOD network, there is usually no box regression part, while this branch plays an important role in fully supervised object detection networks. The box regression network can adjust the position and scale of the proposal, make it closer to the ground truth. In the fully supervised object detection task, we can use instance-level label as supervision to train the box regression network; but in the WSOD task, the network cannot obtain the instance-level annotation, and thus cannot train this branch. In order to obtain reliable instance annotations to train the regression network, we designed a proposal scoring network named PSNet that can detect whether the proposal completely covers the object. The PSNet is a specially trained multi-label classification network. Even if an object in the image is occluded or incomplete, the PSNet can detect the presence of the object. The PSNet can be used to evaluate image without the proposal area. If the proposal completely covers the whole object, the rest of the image will not contain information about it. We use the PSNet to evaluate the output of the WSOD network, and then select the appropriate proposals as pseudo labels to train box regression network. Examples of the output of PSNet are shown in Figure \ref{fig:figure1}.

To verify the effectiveness of our approach, we conducted experiments on the VOC2007 dataset. The end result is xx.

The contributions of the paper are as followings:

1.Introduce box regression branch to WSOD network, reduce the difference between proposal and ground truth.

2.Propose the PSNet to evaluate proposals, as a supervision training box regression branch.

3.Our work is simple to apply and does not require modifications to the backbone. Most end-to-end WSOD network can be used as our backbone.

\section{Related Works}

Weakly supervised object detection task use only image-level annotations, which is an effective way to utilize datasets without instance-level labels. In recent years, there have been a lot of work for this task, which can be roughly divided into three approaches: the alternating approach, end-to-end approach, and two stage approach.

\subsection{Alternating approach}
The early weakly supervised object detection work uses an alternate approach, in which iteratively train the detector and update pseudo labels, so that an effective detector could be trained from the initial rough annotations. A typical alternative approach is~\cite{Song2014On}, in which song et al. assumed that the object exists at the center of the image and trained a detector with such unreliable annotations. Obviously, such a hypothesis is unfounded, and the detectors trained with these unreliable annotations has a poor performance. Song utilized the detector to generate inference in the training dataset, obtained new instance labels. Unreliable annotations can be updated with these instance labels. Repeating the processes that training the detector and update the instance annotations resulted in a stable result.

Some other alternate methods are: Cinbis et al.~\cite{cinbis2016weakly} proposed a multi-fold learning method to solve the problem that alternating approaches are easily trapped in local optima. Li et al. ~\cite{li2016weakly} did not iteratively update detector and pseudo labels, but trained a classifier with the entire image, and then used the mask-out strategy to select the most confident proposal from the feature map of the classifier. Jie et al.~\cite{jie2017deep} proposed a self-taught learning method to select reliable seed proposals. However, the current series of alternative methods are not performing well, because alternating approaches are time-consuming and easily trapped in local optima.

\subsection{End-to-end approach}
The end-to-end approach does not iteratively update the detector and pseudo-annotations, but instead uses an image level annotation to train an end-to-end network. Early end-to-end networks skillfully used proposals to generate the classification outputs of images and trained the network with classification losses. Bilen et al.~\cite{bilen2016weakly} proposed an end-to-end method called weakly supervised deep detection network (WSDDN). The WSDDN has two streams, a classification stream and a detection stream. The results of these two streams are combined to determine the detection score of each proposal and the classification confidence of the image. Kantorov et al.~\cite{kantorov2016contextlocnet} extended WSDDN to utilize contextual information. Diba et al.~\cite{diba2017weakly} and Wei et al.~\cite{wei2018ts2c} take advantage of semantic segmentation network and CAM\cite{zhou2016learning} to select region proposals that tightly cover the instance. Kosugi et al.~\cite{kosugi2019object} developed a context classification network to select the proposal that cover exactly the whole instance in image, and proposed CAP Labeling and SRN Labeling methods to improve the performance of OICR. Tang et Al.~\cite{tang2018weakly} developed high-accuracy region proposals by exploiting the low-level information in CNN.

Tang also proposed two other end-to-end networks, OICR~\cite{tang2017multiple} and PCL~\cite{tang2018pcl}. OICR added iterative instance classifier stream to the WSDDN structure. The instance classifier stream takes the output of previous instance classifier network as supervision to train the next instance classifier. OICR incorporates the idea of the alternate method into an end-to-end approach, generates a more accurate proposal by combining the results of multiple streams. The PCL network is also an end-to-end network based on OICR that take advantage of cluster method. In PCL the spatially overlapping proposals are grouped into one set. The proposals in the same set are more likely to be different parts of the same instance, and the information of multiple proposals can be combined to find the most appropriate proposal. Because of the good performance of the PCL network, our work takes it as the baseline.

\subsection{Two stage approach}
The main feature of the two-stage method is to use other methods to generate pseudo-labels, and then take a full-supervised method to train an object detection network. The first phase of the two-stage approach is to generate pseudo labels by other methods, so it often incorporates other end-to-end methods or alternate methods.

Zhang et al.~\cite{zhang2018w2f} took advantage of WSDDN to propose inaccurate proposals in the image, and took the PGA and PGE two proposal fusion algorithm to get a more accurate proposal to train the faster R-CNN~\cite{ren2015faster} model. Zhang et al.~\cite{zhang2018zigzag} proposed an algorithm named mEAS(mean Energy Accumulated Scores) to calculate the complexity of image content. The training dataset is divided into different groups according to different content complexity, and the fully supervised object detection network is trained with different groups by turns according to the difficulty on the basis of the original two-stage method. In this case, the model can start training from a simpler task and prepare for later difficult training goals, so as to obtain better detection results. Zeng et al.~\cite{zeng2019wsod2} revised the shape of the proposals using the low-level features such as superpixel segmentation, boundary, texture color, etc. The two-stage approach can make full use of the end-to-end network results and facilitate the integration of various rule-based proposal update methods, so there have been many new work in recent years.

\subsection{Bounding box regression}
Bounding box regression can reduce localization errors of predicted boxes, which is first introduced in~\cite{girshick2014rich}. Fully supervised object detection network takes it to get more accurate results, while the box regression module is rare in WSOD network due to the lack of supervision. Gao et al.~\cite{gao2018c} introduced bounding box regressors into OICR network to help selecting pseudo ground truths. Zeng et al.~\cite{zeng2019wsod2} took advantage of the fusion of bottom-up feature and top-down feature as supervision to train box regressors. In our paper, we propose a proposal scoring network(PSNet) to evaluate the appearance of proposals, which can be a supervision of box regressor.

\section{Method}
The backbone of our work is PCL~\cite{tang2018pcl}, which is an end-to-end weak supervised object detection network. The network structure of PCL is similar to that of OICR, except that PCL owns a proposal cluster method.

The PCL network consists of two modules: Basic MIL network and instance refinement module. The basic MIL network is WSDDN~\cite{bilen2016weakly}, which is a WSOD network of two streams. The instance refinement module is composed of multi instance classifier refinement networks. Every instance classifier refinement network contains a fully connected layer and a softmax layer.

The PCL network take advantage of the WSDDN to generate the initial object detection results. The basic MIL network WSDDN consists of two streams, a classification stream and a detection stream, that calculate region-wise scores in a different way based on CNN features pooled by Spatial Pyramid Pooling (SPP)~\cite{He2014Spatial}. The classification stream applies softmax operation on proposals to get classification output. The detection stream applies the softmax operation on the classes to get the contribution of different proposals to different classes. The final object detection results are elementwise product of outputs of the two streams.

The object detection results initialized by WSDDN can be viewed as pseudo labels to train instance refinement module. The instance refinement module contains multiple instance classifier refinement networks with the same structure. These streams are connected one by one, which means that the output of the previous instance classifier refinement network is taken as the pseudo labels to train the next instance classifier refinement network, and the first instance classifier refinement network take outputs from the WSDDN as pseudo labels. Compared with OICR, PCL apply the proposal cluster method to the process of training the instance refinement module, and enhance the performance of pseudo label by clustering the proposals that may belong to the same instance.

\subsection{PSNet}
In the previous WSOD network, the output proposals of the network are not always close to the ground truth. In many cases, the proposal with the highest score contains only the most discriminative area of the object, or contains too many unnecessary background areas. The most important reason for this is that the detector-based method does not use the context information of the proposal, and only determines whether the proposal contains the special object. This will cause some of the proposals that are much smaller than the ground truth to be higher scored than that closer to the ground truth; and some proposals that contain too many background areas will have high scores because they contain some area of the object. Obviously we need to solve the above two problem.

In fact, a good proposal with high score should have the following two features:

Completeness: the proposal should completely contain all the pixels of an instance.

Compactness: the proposal should not contain unnecessary background areas.

In order to judge whether a proposal is good, we propose a completeness detection network named PSNet. The network detects whether the image contains an area of an object. We fill the proposal area in the image with the mean pixel, and then put the whole image into the PSNet. The more object areas contained in the processed image, the higher the output score of the network. We can judge the completeness and compactness of the proposal based on the PSNet output score.

Our proposed PSNet is a multi-label classification network with 21 outputs. An effective multi-label classification network can determine whether there is an object of specific category in an image. But there are two problems if we directly using the multi-label classification network as a proposal scoring network: 1. The network can't distinguish between the object and its environment, especially those with a fixed background. For example, even if there is no train in the image, as long as the railway appears, the network will judge that there is a train, because the two are related to each other. 2. The network can't work well when there are multiple objects of the same category in the image. When there are multiple instances of the same class, even if a proposal completely covers one of those, the network will find the remaining instances, and the classification accuracy will not be significantly reduced.

\begin{figure}[t]
	\begin{center}
		\includegraphics[width=0.9\linewidth]{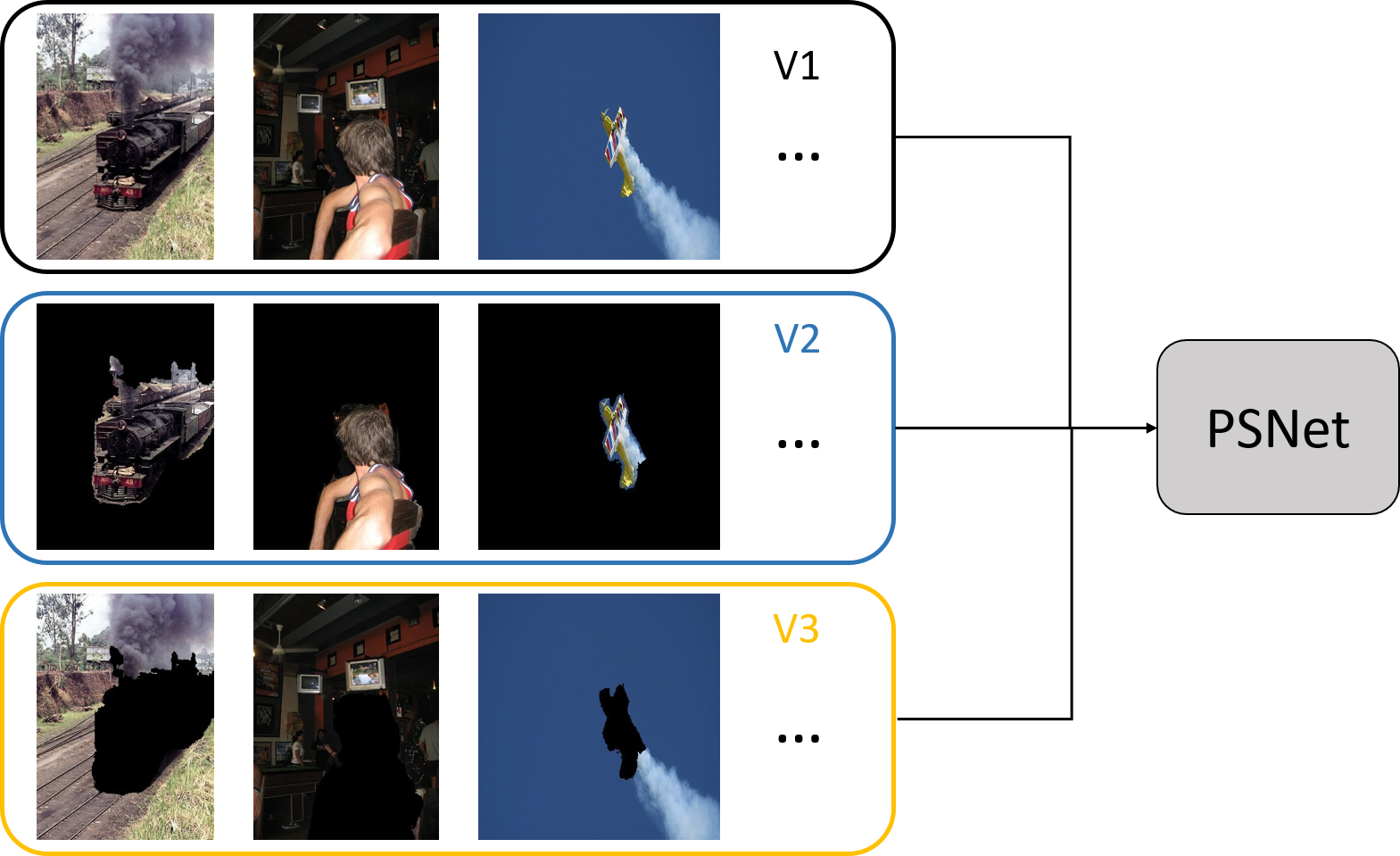}
	\end{center}
	\caption{The PSNet training process is schematically illustrated, and V1, V2, and V3 are shown in the figure. Train with each dataset by turn.}
	\label{fig:figure3}
\end{figure}

To solve these two problems, we use a class-agnostic saliency detection network. The saliency detection network can detect areas of strong saliency in the image. The strong saliency areas are usually objects in the image, and the weak saliency areas are usually the background. One simple way to use the saliency images is to remove all background areas in the datasets and only use the images with foreground area to train the PSNet. However, the effect of this method is not good, and the trained PSNet has a significant drop in classification accuracy on the test set. As shown in Figure \ref{fig:figure3}, the labeled instances in the image are not salient objects, and the foreground area detected by the saliency detection network does not include these instances, which generates noisy labels. Because of these noisy labels, the network accuracy will drop dramatically.

We take advantage of the VOC2007 train dataset and saliency detection network to create three datasets: V1, V2, and V3, and use the three datasets to train the PSNet in order to avoid the effects of noisy annotations. The V1 dataset is the original VOC2007 train dataset. The V2 dataset is the foreground of the V1 dataset segmented by the saliency detection network, and the V3 dataset is the background of the V1 dataset segmented by the saliency detection network. The V1 dataset ensures that the classification network has reliable training labels. The V2 and V3 datasets enable the network to decouple objects from their background, avoiding the misidentification of the environment as objects. The PSNet trained by the above method can effectively solve the first problem.

The saliency detection network can also solve the second problem: when there are multiple instances of the same category in an image, hiding an instance has little effect on the classification accuracy. If the instances of the same category are not close together, their saliency segmentation areas are non-connected regions. So when we use PSNet to test the completeness of a proposal, we only keep the foreground area with the highest IoU and fill the other foreground areas with the mean pixels. This method can reduce the influence of different instances in the picture where the spatial location is far away, but there is no effect on the influence between similar instances in the picture where the spatial location is very close.

In order to separate different instances whose salient areas connected, we used a seed region method. First, we apply different thresholds into the results of the saliency detection network, then obtain different segmented images S1(with high threshold) and S2(with low threshold). The area connected in S2 is dispersed into several non-connected areas in S1, and these non-connected areas are named seed areas. These seed areas in S1 can grow and expand outward until it eventually fills up the saliency area in S2. The above operation can obtain independent saliency region of different instances, separate different instances that are close to but not overlapped, effectively solving the problem of interference of multiple instances.

\begin{figure}[t]
	\begin{center}
		\includegraphics[width=0.9\linewidth]{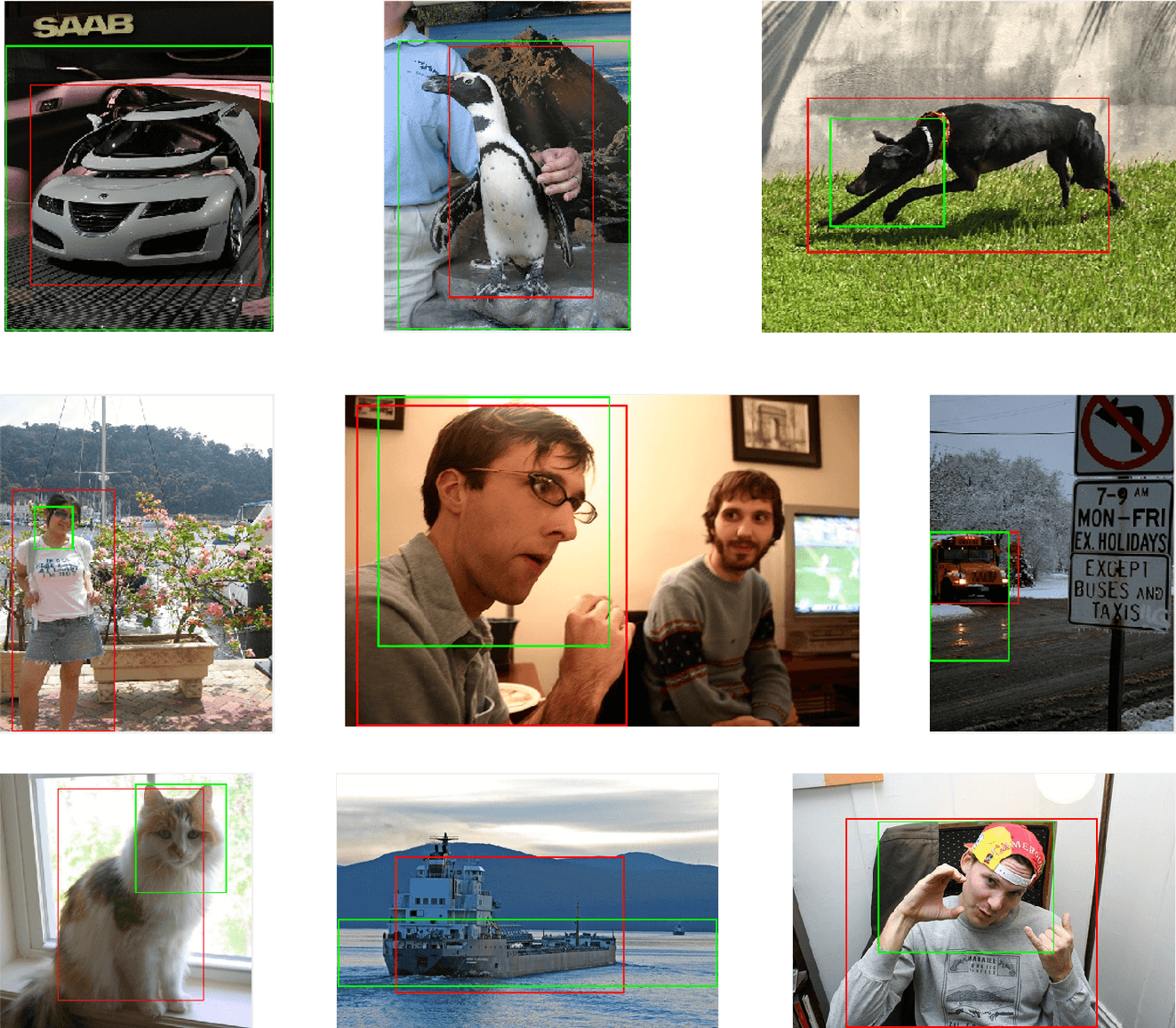}
	\end{center}
	\caption{Some results of PRS algorithm: the green box in the figure represents the proposal found by the PCL~\cite{tang2018pcl} network, and the red box represents the proposal obtained after the PRS algorithm. Obviously our approach can get a proposal that closer to the ground truth.}
	\label{fig:figure4}
\end{figure}

\subsection{Box regression}
In our work, we add the box regression branch to the existing WSOD network PCL. The role of this branch is to revise the proposal that differs from ground truth, making it more accurate. The input of this branch is features extracted from the proposals through the VGG16~\cite{simonyan2014very} conv module, and the output is a prediction of the difference between the proposals and the closest ground truth. In our paper, the input is a feature vector of N*4096??? dimension, and the output is a predicted vector of N*4 dimension, where N is the number of proposals.

Training a box regression network requires instance-level annotation, which is difficult to obtain in weakly supervised object detection task. Thus in the previous WSOD work, the proposal is mostly unchanged. The mainly task of these network is to select the proposal that is most likely to contain instance from a series of proposals. To change this condition, we propose the PSNet network that can score the accuracy of the proposal. If the difference between the proposal and the ground truth is small, the output of the PSNet is very close to 0, and vice versa. Therefore, the output of PSNet can be used as a loss to train box regression network.

However, the process in which the outputs of the box regression network are converted to the input of PSNet is a non-mathematical process, as showing in Figure \ref{fig:figure2}. Because the operation that fill the inside of the proposal with the mean pixels cannot be expressed by a mathematical formula, and it cannot be derived. This means the loss backward will be truncated at the input of the PSNet, so the loss cannot be passed to the parameters of the box regression module. Therefore, we take advantage of an iterative optimization approach to utilize PSNet to train the regression network. The detailed steps are: 1. Apply random offset to the output of the box regression network to get 15 different proposals. 2. Send them together with the original proposal into the PSNet network. 3. Select the highest-scored proposal and then apply random offset to it to get the new 15 proposals. 4. Repeat steps 2 and 3 until the highest scored proposal stable. 5.The final proposal obtained in step 4 is used as a pseudo-label for training the regression network.

We take this formula to score proposals in the second step:
\begin{equation}\label{equ:Sproposal}S_{proposal} = S_{PSNet} + 0.001*S_{area}\end{equation}

The higher the $S_{proposal}$, the better the proposal. $S_{PSNet}$ and $S_{area}$ in (\ref{equ:Sproposal}) are defined as:

\begin{equation}S_{PSNet} = PSNet_{oral} – PSNet_{new}\end{equation}

$PSNet_{oral}$ is the classification output of PSNet for the original proposal, and $PSNet_{new}$ is the classification output of PSNet for the new proposal.

\begin{equation}S_{area}=\begin{cases}
-\log_2(\frac{Area2}{Area1})&\text{$Area2>Area1$},\\
1-\frac{Area2}{Area1}&
\text{$Area2<Area1$}.
\end{cases}\end{equation}

$Area1$ is the area of the original proposal, and $Area2$ is the area of the new proposal.

\section{Experiments}
We experimented on PASCAL VOC dataset to verify the performance of our work.

\subsection{Datasets and parameters}
We conducted experiments on the PASCAL VOC2007 and 2012 datasets. The VOC2007 dataset contains 20 classes and 9,962 images (5011 for training and 4951 for testing). The VOC2007+2012 dataset has 20 classes, including 22,531 images (11540 for training and 10991 for testing). The images in the PASCOL VOC dataset are divided into three groups: train, test, and trainval for training, testing, and cross-validation. We employ mAP and CorLoc to measure the performance of our network. mAP measures the performance of the network on the test set. We take an IoU threshold of 0.5. CorLoc measures the localization accuracy on the trainval dataset

\begin{table*}
	\footnotesize
	\centering
	\renewcommand\tabcolsep{2.9pt}
	\renewcommand\arraystretch{1.5}
	\caption{Average precision (\%) on PASCAL VOC 2007.}
	\begin{center}
		\begin{tabular}{l|cccccccccccccccccccc|c}
			\hline
			method & aero & bike & bird & boat & bottle & bus & car & cat & chair & cow & table & dog & horse & mbike & person & plant & sheep & sofa & train & tv & mAP\\
			\hline\hline
			- VOC 2007&&&&&&&&&&&&&&&&&&&&\\
			OICR~\cite{tang2017multiple} & 58.0 & 62.4& 31.1& 19.4& 13.0& 65.1& 62.2& 28.4& 24.8& 44.7& 30.6& 25.3& 37.8& 65.5& 15.7& 24.1& 41.7& 46.9& 64.3& 62.6&41.2\\
			SGWSOD~\cite{lai2017saliency} & 48.4 & 61.5 & 33.3 & 30.0 & 15.3 & 72.4 & 62.4 & 59.1 & 10.9 & 42.3 & 34.3 & 53.1 & 48.4 & 65.0 & 20.5 & 16.6 & 40.6 & 46.5 & 54.6 & 55.1 & 43.5 \\
			TS2C~\cite{wei2018ts2c} & 59.3 & 57.5 & 43.7 & 27.3 & 13.5 & 63.9 & 61.7 & 59.9 & 24.1 & 46.9 & 36.7 & 45.6 & 39.9 & 62.6 & 10.3 & 23.6 & 41.7 & 52.4 & 58.7 & 56.6 & 44.3\\
			WSRPN~\cite{tang2018weakly} & 57.9 & 70.5 & 37.8 & 5.7 & 21.0 & 66.1 & 69.2 & 59.4 & 3.4 & 57.1 & 57.3 & 35.2 & 64.2 & 68.6 & 32.8 & 28.6 & 50.8 & 49.5 & 41.1 & 30.0 & 45.3\\
			PCL~\cite{tang2018pcl} & 57.1 & 67.1 & 40.9 & 16.9 & 18.8 & 65.1 & 63.7 & 45.3 & 17.0 & 56.7 & 48.9 & 33.2 & 54.4 & 68.3 & 16.8 & 25.7 & 45.8 & 52.2 & 59.1 & 62.0 & 45.8\\
			Kosugi et al.~\cite{kosugi2019object}& 61.5 & 64.8 & 43.7 & 26.4 & 17.1 & 67.4 & 62.4 & 67.8 & 25.4 & 51.0 & 33.7 & 47.6 & 51.2 & 65.2 & 19.3 & 24.4 & 44.6 & 54.1 & 65.6 & 59.5 & 47.6\\
			\hline
			Ours & 62.1 & 67.9 & 51.7 & 22.3 & 18.4 & 69.3 & 68.0 & 47.9 & 23.1 & 54.9 & 42.2 & 49.0 & 51.3 & 67.3 & 13.0 & 24.0 & 46.6 & 53.1 & 61.8 & 58.9 & 47.6\\
			Ours+PRS & 62.9 & 67.8 & 51.7 & 22.3 & 21.0 & 69.3 & 68.1 & 57.7 & 23.1 & 54.8 & 42.2 & 52.7 & 52.2 & 67.1 & 17.5 & 24.1 & 46.6 & 53.7 & 62.3 & 58.9 & 48.8\\
			\hline
		\end{tabular}
	\end{center}
\end{table*}

\begin{table*}
	\footnotesize
	\centering
	\renewcommand\tabcolsep{2.9pt}
	\renewcommand\arraystretch{1.5}
	\caption{CorLoc (\%) on PASCAL VOC 2007 datasets.}
	\begin{center}
		\begin{tabular}{l|cccccccccccccccccccc|c}
			\hline
			method & aero & bike & bird & boat & bottle & bus & car & cat & chair & cow & table & dog & horse & mbike & person & plant & sheep & sofa & train & tv & mean\\
			\hline\hline
			- VOC 2007&&&&&&&&&&&&&&&&&&&&\\
			OICR~\cite{tang2017multiple} & 81.7 & 80.4 & 48.7 & 49.5 & 32.8 & 81.7 & 85.4 & 40.1 & 40.6 & 79.5 & 35.7 & 33.7 & 60.5 & 88.8 & 21.8 & 57.9 & 76.3 & 59.9 & 75.3 & 81.4 & 60.6\\
			TS2C~\cite{wei2018ts2c} & 84.2 & 74.1 & 61.3 & 52.1 & 32.1 & 76.7 & 82.9 & 66.6 & 42.3 & 70.6 & 39.5 & 57.0 & 61.2 & 88.4 & 9.3 & 54.6 & 72.2 & 60.0 & 65.0 & 70.3 & 61.0\\
			SGWSOD~\cite{lai2017saliency} & 71.0 & 76.5 & 54.9 & 49.7 & 54.1 & 78.0 & 87.4 & 68.8 & 32.4 & 75.2 & 29.5 & 58.0 & 67.3 & 84.5 & 41.5 & 49.0 & 78.1 & 60.3 & 62.8 & 78.9 & 62.9\\
			PCL~\cite{tang2018pcl} & 81.7 & 82.4 & 63.4 & 41.0 & 42.4 & 79.7 & 84.2 & 54.9 & 23.4 & 78.8 & 54.4 & 46.0 & 75.9 & 89.6 & 22.8 & 51.3 & 72.2 & 66.1 & 74.9 & 76.0 & 63.0\\
			WSRPN~\cite{tang2018weakly} & 77.5 & 81.2 & 55.3 & 19.7 & 44.3 & 80.2 & 86.6 & 69.5 & 10.1 & 87.7 & 68.4 & 52.1 & 84.4 & 91.6 & 57.4 & 63.4 & 77.3 & 58.1 & 57.0 & 53.8 & 63.8\\
			Kosugi et al.~\cite{kosugi2019object}& 85.5 & 79.6 & 68.1 & 55.1 & 33.6 & 83.5 & 83.1 & 78.5 & 42.7 & 79.8 & 37.8 & 61.5 & 74.4 & 88.6 & 32.6 & 55.7 & 77.9 & 63.7 & 78.4 & 74.1 & 66.7\\
			\hline
			Ours & 82.1 & 75.7 & 73.0 & 43.1 & 43.5 & 76.7 & 83.6 & 65.4 & 40.7 & 76.7 & 44.5 & 62.3 & 77.9 & 88.0 & 36.5 & 54.6 & 65.0 & 59.1 & 74.9 & 74.2 & 64.9\\
			\hline
		\end{tabular}
	\end{center}
\end{table*}

\subsection{Implementation}
Our PSNet take VGG16~\cite{simonyan2014very} as the backbone. The complete network structure: remove the VGG16 classifier module and add a average pooling layer and two fully connected layers. The last fully connected layer has 21 outputs (20 classes + background). We use the convolutional layer parameters of the pre-trained VGG16 on the imagenet as feature extractors. During training process, we keep the convolution module parameters of the network unchanged. The Loss function is BCELoss, using the SGD optimizer, a total of 15 epoch training. The first ten epochs have a lr of 1e-2 and a momentum of 0.9; the last five epochs have a lr of 1e-3 and a momentum of 0.9.

The box regression module is composed of three fully connected layers fc1, fc2, and fc3. The fc1 and fc2 are the same as fc6 and fc7 in VGG16 network. The input of the fc3 layer is 4096, and output is 4-dimensional(tx,ty,tw,th). We initialized fc1 and fc2 with VGG16 pretrained parameters on imagenet, and initialized fc3 randomly. During training, the Loss function is SmoothL1Loss, using the SGD method, a total of 8 epoch. The initial lr is 1e-3, and after every two epochs, the lr is reduced to 10 percent of previous. The momentum is 0.9.

\section{Conclusions}








\section*{Reference}
\bibliography{mybibfile}
\bibliographystyle{elsarticle-num}

\end{document}